\documentclass{esannV2}
\usepackage{graphicx}
\usepackage[utf8]{inputenc}
\usepackage{amssymb,amsmath,array}
\usepackage[english]{babel}
\usepackage{microtype}
\usepackage{url}

\begin{document}
\title{Using the Mean Absolute Percentage Error for Regression Models}

\author{Arnaud de Myttenaere$^{1,2}$, Boris Golden$^1$, B\'en\'edicte Le Grand$^3$ \& Fabrice Rossi$^2$
\vspace{.3cm}\\
1 - Viadeo\\
30 rue de la Victoire, 75009 Paris - France
\vspace{.1cm}\\
2 - Universit\'e Paris 1 Panth\'eon - Sorbonne - SAMM EA 4534 \\
90 rue de Tolbiac, 75013 Paris - France
\vspace{.1cm}\\
3 - Universit\'e Paris 1 Panth\'eon - Sorbonne - Centre de Recherche en Informatique \\
90 rue de Tolbiac, 75013 Paris - France\\
}

\maketitle

\begin{abstract}
We study in this paper the consequences of using the Mean Absolute Percentage
Error (MAPE) as a measure of quality for regression models. We show that finding the
best model under the MAPE is equivalent to doing weighted Mean Absolute
Error (MAE) regression. We show that universal consistency of Empirical Risk Minimization
remains possible using the MAPE instead of the MAE. 
\end{abstract}

\section{Introduction}\label{sec:introduction}
We study in this paper the classical regression setting in which we assume
given a random pair $Z=(X,Y)$ with values in $\mathcal{X}\times\mathbb{R}$,
where $\mathcal{X}$ is a metric space. The goal is to learn a mapping $g$ from
$\mathcal{X}$ to $\mathbb{R}$ such that $g(X)\simeq Y$. To judge the quality
of the regression model $g$, we need a quality measure. While the traditional
measure is the quadratic error, in some applications, a more useful measure of
the quality of the predictions made by a regression model is given by the mean
absolute percentage error (MAPE). For a target $y$ and a prediction $p$, the
MAPE is 
\[
l_{MAPE}(p,y)=\frac{|p-y|}{|y|},
\]
with the conventions that for all $a\neq 0$, $\frac{a}{0}=\infty$ and that
$\frac{0}{0}=1$. The MAPE-risk of $g$ is then
$L_{MAPE}(g)=\mathbb{E}(l_{MAPE}(g(X),Y))$. 

We are interested in the consequences of choosing the best regression model
according to the MAPE as opposed to the Mean Absolute Error (MAE) or the
Mean Square Error (MSE), both on a practical point of view and on a
theoretical one. On a practical point of view, it seems obvious that if $g$ is
chosen so as to minimize $L_{MAPE}(g)$ it will perform better according to
the MAPE than a model selected in order to minimize $L_{MSE}$ (and worse
according to the MSE). The practical issue is rather to determine how to
perform this optimization: this is studied in Section \ref{sec:practical}. On
a theoretical point of view, it is well known 
(see e.g. \cite{gyorfi_etal_DFTNR2002}) that consistent learning schemes can
be obtained by adapting the complexity of the model class to the data size. As
the complexity of a class of models is partially dependent on the loss
function, using the MAPE instead of e.g. the MSE has some implications that
are investigated in this paper, in Section \ref{sec:theory}. The following
Section introduces the material common to both parts of the analysis. 

\section{General setting}\label{sec:general-setting}
We use a classical statistical learning setting as in
e.g. \cite{gyorfi_etal_DFTNR2002}. We assume given $N$ independently
distributed copies of $Z$, the training set, $D=(Z_i)_{1\leq i\leq
  N}=(X_i,Y_i)_{1\leq i\leq N}$. Given a loss function $l$ from $\mathbb{R}^2$
to $\mathbb{R}^+\cup\{\infty\}$, we define the risk of a predictor $g$, a
(measurable) function from $\mathcal{X}$ to $\mathbb{R}$ as the expected loss,
that is $L_l(g)=\mathbb{E}(l(g(X),Y))$. The empirical risk is the empirical
mean of the loss computed on the training set, that is:
\begin{equation}\label{eq:empirical:risk}
\widehat{L}_l(g)_N=\frac{1}{N}\sum_{i=1}^Nl(g(X_i),Y_i).
\end{equation}
In addition to $l_{MAPE}$ defined in the Introduction, we use
$l_{MAE}(p,y)=|p-y|$ and $l_{MSE}(p,y)=(p-y)^2$. 

\section{Practical issues}\label{sec:practical}
\subsection{Optimization}
On a practical point of view, the problem is to minimize
$\widehat{L}_{MAPE}(g)_N$ over a class of models $G_N$, that is to
solve\footnote{We are considering here the empirical risk minimization, but we
could of course include a regularization term. That would not modify the key
point which is the use of the MAPE.}
\[
\widehat{g}_{MAPE,N}=\arg\min_{g\in G_N}\frac{1}{N}\sum_{i=1}^N\frac{|g(X_i)-Y_i|}{|Y_i|}.
\]
Optimization wise, this is simply a particular case of \emph{median regression}
(which is in turn a particular case of \emph{quantile regression}). Indeed,
the quotient by $\frac{1}{|Y_i|}$ can be seen as a fixed weight and therefore,
any quantile regression implementation that supports instance weights can be
use to find the optimal model\footnote{This is the case of
\texttt{quantreg} R package \cite{Quantreg}, among others.}. Notice that when $G_N$
corresponds to linear models, the optimization problem is a simple
\emph{linear programming} problem that can be solved by e.g. interior point
methods~\cite{BoydVandenbergheCVBook2004}. 

\subsection{An example of typical results}
We verified on a toy example (the car data set from \cite{cars1930dataset})
the effects of optimizing the MAPE, the MAE and the MSE for a simple linear
model: the goal is to predict the distance taken to stop from the speed of the
car just before breaking. There are only 50 observations, the goal being here
to illustrate the effects of changing the loss function. The results on the
training set\footnote{Notice that the goal here is to verify the effects of
  optimizing with respect to different types of loss function, not to claim
  that one loss function is better than another, something that would be
  meaningless. We report therefore the empirical risk, knowing that it is an
  underestimation of the real risk for all loss functions.} are summarized in
Table \ref{tab:results}. As expected, optimizing for a particular loss
function leads to the best empirical model as measured via the same 
risk (or a related one). In practice this allowed one of us to win a recent \url{datascience.net}
challenge about electricity consumption
prediction\footnote{\url{https://datascience.net/fr/challenge/16/details}}
which was using the MAPE as the evaluation metric.

\begin{table}[htbp]
  \centering
  \begin{tabular}{lccc}
Loss function & RMSE & NMAE & MAPE\\\hline
MSE & \textbf{0.585} & 0.322 & 0.384 \\
MAE & 0.601 & \textbf{0.313} & 0.330\\
MAPE& 0.700 & 0.343 & \textbf{0.303}\\\hline
  \end{tabular}
\vspace{-0.5em}
  \caption{Empirical risks of the best linear models obtained with the three loss functions. In order to ease the comparisons between the values, we report the Normalized Root MSE, that is the square root of the MSE divided by the standard deviation of the target variable, as well as the Normalized MAE, that is the MAE devided by the median of the target variable.}
  \label{tab:results}
\end{table}
\vspace{-2em}

\section{Theoretical issues}\label{sec:theory}
On a theoretical point of view, we are interested in the consistency of
standard learning strategies when the loss function is the MAPE. More
precisely, for a loss function $l$, we define $L^*_l=\inf_{g}L_l(g)$, where
the infimum is taken over all measurable functions from $\mathcal{X}$ to
$\mathbb{R}$. We also denote $L^*_{l,G}=\inf_{g\in G}L_l(g)$ where $G$ is a
class of models. Then a learning algorithm, that is a function which maps the
training set $D=(X_i,Y_i)_{1\leq i\leq N}$ to a model $\widehat{g}_N$, is strongly
consistent if $L_l(\widehat{g}_N)$ converges almost surely to $L^*_l$. We are
interested specifically by the Empirical Risk Minimization (ERM) algorithm,
that is by $\widehat{g}_{l,N}=\arg\min_{g\in G_N}\widehat{L}_l(g)_N$. The
class of models to depend on the data size as this is 
mandatory to reach consistency. 

It is well known (see e.g. \cite{gyorfi_etal_DFTNR2002} chapter 9) that ERM
consistency is related to uniform laws of large numbers (ULLN). In particular, we
need to control quantities of the following form
\begin{equation}
  \label{eq:uniform}
  P\left\{\sup_{g\in G_N}\left|\widehat{L}_{mape}(g)_N-L_{mape}(g)\right|>\epsilon\right\}.
\end{equation}
This can be done via covering numbers or via the Vapnik-Chervonenkis dimension
(VC-dim) of certain classes of functions derived from $G_N$. One
might think that general results about arbitrary loss functions can be used to
handle the case of the MAPE. This is not the case as those results generally
assume a uniform Lipschitz property of $l$ (see Lemma 17.6 in
\cite{ab-nnltf-99}, for instance) that is not fulfilled by the MAPE.

\subsection{Classes of functions}
Given a class of models, $G_N$, and a loss function $l$, we introduce derived
classes $H(G_N,l)$ given by
\[
H(G_N,l)=\{h: \mathcal{X}\times \mathbb{R}\rightarrow \mathbb{R}^+,\
h(x,y)=l(g(x),y)\ |\ g\in G_N\},
\]
and $H^+(G_N,l)$ given by
\[
H^+(G_N,l)=\{h: \mathcal{X}\times \mathbb{R}\times \mathbb{R}\rightarrow \mathbb{R}^+,\
h(x,y,t)=\mathbb{I}_{t\leq l(g(x),y)}\ |\ g\in G_N\}.
\]
When this is obvious from the context, we abbreviate the notations into e.g.
$H_{N,MAPE}$ for $l=l_{MAPE}$ and for the $G_N$ under study.

\subsection{Covering numbers}
\subsubsection{Supremum covering numbers} 
Let $\epsilon>0$, a size $p$ supremum $\epsilon$-cover of a class of positive functions
$F$ from an arbitrary set $\mathcal{Z}$ to $\mathbb{R}^+$ is a finite collection
$f_1,\ldots,f_p$ of $F$ such that for all $f\in F$
\[
\min_{1\leq i\leq p}\sup_{z\in\mathcal{Z}}|f(z)-f_i(z)|<\epsilon.
\]
Then the supremum $\epsilon$-covering number of $F$,
$\mathcal{N}_{\infty}(\epsilon,F)$, is the size of the smallest supremum
$\epsilon$-cover of $F$. If such a cover does not exists, the covering number
is $\infty$. While controlling supremum covering numbers of $H(G_N,l)$ leads
easily to consistency via a uniform law of large numbers (see e.g. Lemma 9.1
in \cite{gyorfi_etal_DFTNR2002}), they cannot be used with the MAPE without
additional assumptions. Indeed, let $h_1$ and $h_2$ be two functions from
$H_{N,MAPE}$, generated by $g_1$ and $g_2$ in $G_N$. Then
\[
\|h_1-h_2\|_{\infty}=\sup_{(x,y)\in \mathcal{X}\times \mathbb{R}}\frac{||g_1(x)-y|-|g_2(x)-y||}{|y|}. 
\]
In general, this quantity will be unbounded as we cannot control the behavior
of $g(x)$ around $y=0$ (indeed, in the supremum, $x$ and $y$ are independent
and thus unless $G_N$ is very restricted there is always $x$ and
$g_1$ and $g_2$ such that $g_1(x)\neq g_2(x)\neq 0$). Thus we have to assume that there is $\lambda>0$
such that $|Y|\geq \lambda$. This is not needed when using more traditional
loss functions such as the MSE or the MAE. Then we have 
\[
\mathcal{N}_{\infty}(\epsilon,H(G_N,l_{MAPE}))\leq \mathcal{N}_{\infty}(\lambda\epsilon,H(G_N,l_{MAE})).
\]

\subsubsection{$L_p$ covering numbers}
$L_p$ covering numbers are similar to supremum covering numbers but are based
on a different metric on the class of functions $F$ and are data
dependent. Given a data set $D$, we define
\[
\|f_1-f_2\|_{p,D}=\left(\frac{1}{N}\sum_{i=1}^N|f_1(Z_i)-f_2(Z_i)|^p\right)^{\frac{1}{p}},
\]
and derive from this the associated notion of $\epsilon$-cover and of covering
number. It's then easy to show that
\[
\mathcal{N}_{p}(\epsilon,H(G_N,l_{MAPE}),D)\leq \mathcal{N}_{p}(\epsilon\min_{1\leq i\leq N}|Y_i|,H(G_N,l_{MAE}),D).
\]

\subsection{Uniform law of large numbers}
In order to get a ULLN from a covering number of a class $F$, one needs a uniform bound on
$F$. For instance, Theorem 9.1 from \cite{gyorfi_etal_DFTNR2002} assumes that
there is a value $B_F$ such that for all $f\in F$ and all $z\in\mathcal{Z}$, $f(z)\in
[0,B_F]$. With classical loss functions such as MAE and MSE, this is achieved via
upper bounding assumptions on both $G_N$ and on $|Y|$. In the MAPE case, the
bound on $G_N$ is needed but the upper bound on $|Y|$ is replaced by the lower
bound already needed. Let us assume indeed that for all $g\in G_N$,
$\|g\|_{\infty}\leq B_{G_N}$. Then if $|Y|\leq B_Y$, we have
$B_{H(G_N,l_{MAE})}=B_{G_N}+B_Y:=B_{N,MAE}$, while if $|Y|\geq \lambda$, we
have $B_{H(G_N,l_{MAPE})}=1+\frac{B_{G_N}}{\lambda}:=B_{N,MAPE}$. 

Theorem 9.1 from \cite{gyorfi_etal_DFTNR2002} gives then (with $B_{N,l}=B_{H(G_N,l)}$)
\begin{equation}\label{eq:ULLN}
P\left\{\sup_{g\in G_N}\left|\widehat{L}_{l}(g)_N-L_{l}(g)\right|>\epsilon\right\}
\leq 8\mathbb{E}\left(\mathcal{N}_{p}\left(\frac{\epsilon}{8},H(G_N,l),D\right)\right)e^{-\frac{N\epsilon^2}{128B_{N,l}^2}}.
\end{equation}
The expressions of the two bounds above show that $B_Y$ and
$\lambda$ play similar roles on the exponential decrease of the right hand
side bound. Loosening the condition on $Y$ (i.e., taking a large $B_Y$ or a
small $\lambda$) slows down the exponential decrease.

It might seem from the results on the covering numbers that the MAPE suffers
more from the bound needed on $Y$ than e.g. the MAE. This is not the case 
as bounds hypothesis on $F$ are also needed to get finite covering
numbers (see the following section for an example). Then we can consider that
the lower bound on $|Y|$ plays an equivalent role for the MAPE to the one
played by the upper bound on $|Y|$ for the MAE/MSE. 

\subsection{VC-dimension}
A convenient way to bound covering numbers is to use
VC-dimension. Interestingly replacing the MAE by the MAPE cannot increase the
VC-dim of the relevant class of functions. 

Let us indeed consider a set of $k$
points shattered by $H^+(G_N,l_{MAPE})$, $(v_1,\ldots,v_k)$,
$v_j=(x_j,y_j,t_j)$. Then for each $\theta\in\{0,1\}^k$, there is $h_{\theta}\in
H_{N,MAPE}$ such that $\forall j,\ \mathbb{I}_{t\leq
  h_{\theta}(x,y)}(x_j,y_j,t_j)=\theta_j$. Each $h_\theta$ corresponds to
a $g_\theta\in G_N$ and $t\leq h_{\theta}(x,y)\Leftrightarrow t\leq
\frac{|g_{\theta}(x)-y|}{|y|}$. Then the set of $k$ points defined by
$w_j=(z_j,|y_j|t_j)$ is shattered by $H^+(G_N,l_{MAE})$ because the $h'_{\theta}$ associated in
$H_{N,MAE}$ to the $g_{\theta}$ are such that $\forall j,\ \mathbb{I}_{t\leq
  h'_{\theta}(x,y)}(x_j,y_j,|y_j|t_j)=\theta_j$. Therefore 
\[
V_{MAPE}:=VC_{dim}(H^+(G_N,l_{MAPE}))\leq VC_{dim}(H^+(G_N,l_{MAE})):=V_{MAE}.
\]
Using theorem 9.4 from \cite{gyorfi_etal_DFTNR2002}, we can bound the $L^p$
covering number with a VC-dim based value. If $V_{l}=VC_{dim}(H^+(G_N,l))\geq 2$, $p\geq 1$, and
$0<\epsilon<\frac{B_{N,l}}{4}$, then
\begin{equation}\label{eq:covering:VC}
\mathcal{N}_{p}(\epsilon,H(G_N,l),D)\leq 3\left(\frac{2eB_{N,l}^p}{\epsilon^p}\log \frac{3eB_{N,l}^p}{\epsilon^p}\right)^{V_{l}}.
\end{equation}
When this bound is plugged into equation \eqref{eq:ULLN}, it shows the
symmetry between $B_Y$ and $\lambda$ as both appears in the relevant
$B_{N,l}$. 

\subsection{Consistency}
Mimicking Theorem 10.1 from \cite{gyorfi_etal_DFTNR2002}, we can prove a
generic consistency result for MAPE ERM learning. Assume given a series of
classes of models, $(G_n)_{n\geq 1}$ such that $\bigcup_{n\geq 1}G_n$ is dense in
the set of measurable functions from $\mathbb{R}^p$ to $\mathbb{R}$ according
to the $L^1(\mu)$ metric for any probability measure $\mu$. Assume in addition
that each $G_n$ leads to a finite VC-dim $V_n=VC_{dim}(H^+(G_n,l_{MAPE})$ and
that each $G_n$ is uniformly bounded by $B_{G_n}$. Notice that those two
conditions are compatible with the density condition only if
$\lim_{n\rightarrow\infty}v_n=\infty$ and
$\lim_{n\rightarrow\infty}B_{G_n}=\infty$. 

Assume finally that $(X,Y)$ is such as $|Y|\geq \lambda$ (almost
surely) and that $\lim_{n\rightarrow \infty}\frac{v_nB_{G_n}^2\log B_{G_n}}{n}=0$,
then $L_{MAPE}(\widehat{g}_{l_{MAPE},n})$ converges almost surely to
$L^*_{MAPE}$, which shows the consistency of the ERM estimator for the MAPE. 

The proof is based on the classical technique of exponential bounding. Plugin
equation \eqref{eq:covering:VC} into equation \eqref{eq:ULLN} gives a bound on
the deviation between the empirical mean and the expectation of
\[
K(n,\epsilon)=24\left(\frac{16eB_n}{\epsilon}\log\frac{24eB_n}{\epsilon}\right)^{v_n}e^{-\frac{n\epsilon^2}{128B_n^2}},
\]
with $B_n=1+\frac{B_{G_n}}{\lambda}$. Then it is easy to check that the
conditions above guarantee that $\sum_{n\geq 1} K(n,\epsilon)<\infty$ for all
$\epsilon>0$. This is sufficient to show almost sure convergence of
$L_{MAPE}(\widehat{g}_{l_{MAPE},n})-L^*_{MAPE,G_n}$ to 0. The conclusion
follows from the density hypothesis. 

\section{Conclusion}
We have shown that learning under the Mean Absolute Percentage Error is
feasible both on a practical point of view and on a theoretical one. In
application contexts where this error measure is adapted (in general when the
target variable is positive by design and remains quite far away from zero,
e.g. in price prediction for expensive goods), there is therefore no reason to
use the Mean Square Error (or another measure) as a proxy for the MAPE. An
open theoretical question is whether the symmetry between the upper bound on
$|Y|$ for MSE/MAE and the lower bound on $|Y|$ for the MAPE is strong enough
to allow results such as Theorem 10.3 in \cite{gyorfi_etal_DFTNR2002} in which
a truncated estimator is used to lift the bounded hypothesis on $|Y|$. 

\begin{footnotesize}
\bibliographystyle{abbrv}
\bibliography{mape-theory}
 \end{footnotesize}

\end{document}